\documentclass[conference]{IEEEtran}
\IEEEoverridecommandlockouts

\ifCLASSINFOpdf
  \usepackage[pdftex]{graphicx}
\else
\fi

\usepackage{graphicx} 
\usepackage{subcaption}
\usepackage{threeparttable}

\usepackage{stfloats}
\usepackage{tabularx,booktabs}
\usepackage{hyperref}
\usepackage[final,xcolor={divpdf,grey},authormarkup=none]{changes}

\usepackage[T1]{fontenc}

\begin{document}
\title{High-definition event frame generation using SoC FPGA devices}%The challenges of generating high-definition event frames in FPGA devices}

\author{\IEEEauthorblockN{Krzysztof Blachut, Tomasz Kryjak}
\thanks {The work presented in this paper was supported by the programme ``Excellence initiative –- research university'' for the AGH University of Krakow. % and was partly supported by the National Science Centre project no. 2021/41/N/ST6/03915 entitled ``Acceleration of processing event-based visual data with the use of heterogeneous, reprogrammable computing devices''.
The authors would like to thank Piotr Wzorek and Mateusz Wąsala for their help with the project.}
%\IEEEauthorblockA{Embedded Vision Systems Group\\Computer Vision Laboratory\\ Department of Automatic Control and Robotics\\AGH University of Krakow\\Al. Mickiewicza 30\\30-059 Krakow, Poland\\E-mail: kblachut@agh.edu.pl\\{*}Corresponding author}
\IEEEauthorblockA{
\textit{Embedded Vision Systems Group, Department of Automatic Control and Robotics,} 
\textit{AGH University of Krakow, Poland}}
\textit{\href{mailto:kblachut@agh.edu.pl}{kblachut@agh.edu.pl}, \href{mailto:tomasz.kryjak@agh.edu.pl}{tomasz.kryjak@agh.edu.pl}}
}
\maketitle

\begin{abstract}
%\boldmath
%\deleted{Event cameras are state-of-the-art sensors that only record changes in brightness that occur in the scene being monitored.
%As a~result, they work correctly in a~variety of lighting conditions and have low latency.
%However, due to the different data format, typical vision algorithms cannot be easily applied to them.
%One solution to this problem is the generation of event frames onto which incoming events within a~specified time interval are projected.
%Field Programmable Gate Array (FPGA) devices are a~good platform for the implementation of many vision systems, also working in real-time operation and characterised by low-power consumption.}
In this paper we have addressed the implementation of the accumulation and projection of high-resolution event data \added{stream} (HD -- $1280 \times 720$ pixels) onto the image plane in FPGA devices. %\added{SoC} FPGA \added{(System-on-Chip Field Programmable Gate Array) devices}.
The results confirm the feasibility of this approach, but there are a~number of challenges, limitations and trade-offs to be considered. %as described in this work.
The required hardware resources of selected data representations, such as binary frame, event frame, exponentially decaying time surface and event frequency, were compared with those available on several popular platforms from AMD Xilinx.
The resulting event frames can be used for typical vision algorithms, such as object classification and detection, using both classical and deep neural network methods.
\end{abstract}

% \fi
\IEEEpeerreviewmaketitle

\section{Introduction}
\label{sec:intro}

%stare
%Współczesny świat charakteryzuje się bardzo dużą dynamiką.
%Spostrzeżenie to dotyczy nie tylko ludzi, ale także pojazdów autonomicznych, od których oczekuje się maksymalnie szybkiego, poprawnego i bezpiecznego działania.
%Szczególnie ważne jest to w przypadku bezzałogowych statków powietrznych, zwanych popularnie dronami, które w ostatnich latach znajdują zastosowanie w wielu dziedzinach życia, w tym %chociażby do transportu przesyłek, inspekcji sieci przesyłowych czy pomocy w misjach ratunkowych \cite{Shakhatreh2019}.
%Aby umożliwić większą niezależność od operatora drona i jego umiejętności, coraz częściej dokonuje się autonomizacji części wykonywanych przez drona zadań.
%Można tu wymienić przykładowo detekcję obiektów \cite{Wan2021} lub ich śledzenie, wyznaczanie przepływu optycznego \cite{Liu2017}, czy detekcję i dopasowanie punktów charakterystycznych %\cite{Clady2015}.
%stare
%Sprawne wykonywanie zadań, w tym przykładowo przelot przez dynamiczne środowisko z przeszkodami, wymaga szybkiej akwizycji i przetwarzania danych w celu ominięcia występujących na trajektorii lotu przeszkód.
%Do akwizycji danych najczęściej stosuje się zestaw czujników, takich jak kamery, czujniki inercyjne (IMU –- Inertial Measurement Unit), ultradźwiękowe czy pomiary z systemu GPS.
%Jednak w przypadku dynamicznego ruchu drona lub obiektów w jego otoczeniu są one niewystarczające.

Event cameras (DVS -- Dynamic Vision Sensors) are state-of-the-art vision sensors inspired by biology.
They only detect changes in brightness in an image and can operate with microsecond resolution.
For this reason, they are able to register changes in the environment very quickly, i.e. they have low latency.
In addition, they operate correctly under difficult lighting conditions, e.g. limited brightness or high dynamic range.
The aforementioned properties make DVS an interesting sensor for currently developing autonomous vehicles, including drones, where both speed of operation and robustness to a~variety of lighting conditions are important.

However, processing event data, which is in the form of a~sparse spatio-temporal cloud, is challenging.
Several approaches are encountered in the literature: from the direct analysis of the point cloud, to the projection of events onto a~plane (representations) and reconstruction \cite{Rebecq2021}.
Particularly popular is the use of a~representation in the form of a~so-called event frames, i.e. events collected over a~certain time interval, projected onto the image plane. 
This allows the use of typical, well-known vision algorithms -- both classical and those based on deep learning.

In order to process event data quickly enough, a~powerful hardware platform is needed on which relevant information can be extracted in real-time, e.g. the position and direction of potential objects in the drone's environment.
This prerequisite is met by SoC FPGA (System-on-Chip Field Programmable Gate Array) or embedded GPU (embedded Graphics Processing Unit) platforms, which enable parallel computing and, in addition, are small and lightweight and therefore well suited to be placed on a~drone.

In this paper, we address the \replaced{subject}{topic} of event frame generation in SoC FPGAs for data in HD (High Definition -- $1280 \times 720$ pixels) resolution or higher.
To the best of our knowledge, this issue has not yet been considered in the scientific literature for such high resolutions (Section \ref{ssec:prev_work}) and it presents many challenges, both scientific (hardware architecture development) and technical (implementation).

The remainder of this paper is organised as follows.
Section \ref{sec:dvs} provides background information on event cameras and the approaches available in the scientific literature to the problem \replaced{considered}{under consideration}.
Section \ref{sec:repr} is devoted to the presentation and comparison of different event data representations.
Details of the hardware implementation of the algorithm, together with its potential enhancements, are collected in Section \ref{sec:impl}.
The final Section \ref{sec:concl} summarises the work done and indicates further directions for development.

\section{Event-frame generation}
\label{sec:dvs}

\subsection{Event camera}
\label{ssec:camera}

Dynamic Vision Sensors differ from traditional frame cameras in both design and operation.
Firstly, they only record \added{brightness} changes \deleted{in brightness in the observed scene,} independently (asynchronously) for each pixel. 
Each such change is called an event and consists of four pieces of information: the exact time of occurrence (timestamp), the $x$ and $y$ positions in the image, and the polarity, i.e. information about the increase or decrease in brightness (positive and negative polarity respectively), \replaced{which}{ -- this} can be written as $e = \{t,x,y,p\}$.

When monitoring a~static scene, event cameras do not transmit any data (apart from noise), which significantly reduces data transfer and energy requirements. 
Secondly, changes are recorded at a~very high rate -- more than a~thousand times per second, whereas frame cameras typically operate at \replaced{lower frequencies}{around 60 fps (frames per second)} \added{(e.g. 30, 60, 120 or up to 240 frames per second)}.
Another difference resulting from this is the significant reduction in blur effect associated with object movement on event data, which enables correct registration of a~highly dynamic scene. 
A~fourth advantage over traditional cameras is the high dynamic range, allowing very bright and very dark areas to be recorded in a~single image, which is usually not possible with frame cameras.

Event cameras are used, among others, in unmanned aerial vehicles \cite{Vidal2018}, for increasing the frequency of video sequences \cite{Pan2019}, for traffic sign detection \cite{Wzorek2022}, or for fast objects counting \cite{Bialik2023}.
A~comprehensive overview of event camera applications can be found in the article \cite{Gallego2022}.

\subsection{Previous works}
\label{ssec:prev_work}

A~number of solutions can be found in the scientific literature where frames were generated to perform certain tasks based on event data.
The paper \cite{Wzorek2022} realised traffic sign detection using neural networks and proposed a~multi-channel approach (different \added{simple} representations on three image channels).
The work \cite{Chen2018} proposed a~system for object detection using neural networks and a~representation based on the frequency of event occurrence.
The paper \cite{Wan2021} presented a~system for pedestrian detection using a~neural network and a~representation called neighbourhood suppression time surface.
For the gesture recognition task, a~temporal binary representation method was proposed in \cite{Innocenti2021}, where a~sequence of 8 binary frames was generated, aggregated to a~single greyscale image.
The only work where frame generation was done for HD data is \cite{Bialik2023}, which proposed a~system to control the flow of elements in time (falling corn grains).

One of a~few similar hardware implementations on an FPGA platform is \cite{Linares2021}, whose authors performed object classification using a~NullHop accelerator based on a~convolutional neural network.
In this work, frames were generated for a~specific number of events instead of a~time interval.
Two memory blocks for data of only $64 \times 64$ elements were used, so the authors did not encounter a~number of challenges in implementing the algorithm for high-resolution data.

The generation of frames without distinguishing events' polarity was part of the optical flow determination algorithm on an FPGA platform \cite{Liu2017}, in which data was accumulated in 3 blocks from consecutive time intervals.
In the project an event camera with a~resolution of $240 \times 180$ pixels was used.

The work \cite{Tapiador2020} developed a~gesture recognition system using an FPGA platform and a~hierarchy of time surfaces representation, inspired by an approach from multi-layer neural networks, and a~single-layer linear decay kernel representation, i.e. linear decay of events in time.
Each single event was specified using the time context concept, a~quadratic matrix of timestamp differences in a~small neighbourhood.
In the project two event sensors (with $304 \times 240$ and $128 \times 128$ pixels resolutions) were used, but for the purposes of the algorithm, the data was stored in an accumulator scaled to $128 \times 128$.

In summary, in a~number of research papers \deleted{proposed} projecting event data onto the image plane using different representations \added{was proposed}, and further processing of the frames was \replaced{rather}{usually} performed using neural networks \added{than classical methods}. 
\replaced{In some of these papers, an FPGA platform was used for the implementation of the frame generation algorithm}{Single papers have also been devoted to the implementation of the algorithm on an FPGA platform}, but for very low resolution data.
\deleted{To our knowledge, the problem of frame generation for high-resolution (e.g. HD) event data on an FPGA platform in real-time has not yet been addressed.}
The implementation of such an algorithm \added{for high-resolution data has not yet been addressed as it} raises \deleted{some} \added{several} problems, mainly related to the handling of incoming events and real-time frame generation, which at low resolutions -- due to much less data per time interval -- simply do not occur.
This paper therefore attempts to fill this gap, which becomes particularly relevant as the resolution of commercially available event cameras increases \added{-- from $128 \times 128$ pixels in 2008 up to $1280 \times 960$ in 2020 \cite{Gallego2022}}.

\section{Event-frame representations}
\label{sec:repr}

There are several approaches to processing event data, e.g. `straightforward', which requires the development of new algorithms due to the different data format.
Therefore, a~more `natural' approach is the generation of so-called event frames, which consist of events from certain time intervals projected onto a~plane, resulting in pseudo-images similar to these of traditional frame cameras, to which well-known algorithms can be applied.
However, the frame generation poses a~number of challenges: selecting the event accumulation period, taking into account time information when projecting onto the plane or ensuring real-time frame generation and further processing.

To date, several basic data representations for event frame generation have been proposed in the literature, with additional modifications for specific applications.
%od PW with modifications
According to \cite{Clady2015}, a~function $\Sigma_{e}$ can be defined that assigns the time of occurrence of an event $t$ to each coordinate pair $\textbf{u}(x,y)$, and a~function $P_{e}$ that assigns a~polarity from the set $\{1, -1\}$ or, using another convention, $\{1, 0\}$.
%In the case of the latter function, another convention is also encountered in the literature, in which a value from the set ${1, 0}$ is assigned.
% \begin{equation}
% \label{eq:sigma}
% \textbf{u}: t = \Sigma_{e}(\textbf{u}), \Sigma_{e}: R^{2} \rightarrow R
% \end{equation}
% \begin{equation}
% \label{eq:pe}
% \textbf{u}: p = P_{e}(\textbf{u}), P_{e}: R^{2} \rightarrow \{1, -1\}
% \end{equation}
In addition, a~parameter $\tau$ can be defined, indicating the event accumulation time for the generation of one frame.
The representations listed in this section are defined (for simplicity) for times $t \in (0, \tau)$.

\subsection{Binary frame}
\label{ssec:binary}
The first representation -- for \deleted{reasons of} maximum simplicity -- can be a~binary information about the occurrence of an event in given coordinates, as in \cite{Liu2017}: $f(\textbf{u},t) = 1$.
A~binary frame can also be generated from events of only one polarity. %, e.g. positive.

%PL
% Pierwszym z rodzajów reprezentacji -- z uwagi na maksymalną prostotę -- może być informacja binarna o wystąpieniu zdarzenia dla danych współrzędnych, tak jak w pracy \cite{Liu2017}, zgodnie ze wzorem (\ref{eq:event_bin}).
% Pewnym rodzajem tej reprezentacji może być ramka wygenerowana na podstawie zdarzeń o tylko jednej polaryzacji, np. dodatniej.

% \begin{equation}
% \label{eq:event_bin}
% f(\textbf{u},t) = 1 %\textrm{ for } P_{e}(\textbf{u}) \in \{1, -1\}
% %\left\{ \begin{array}{lr} 255 & \textrm{ for }  P_{e}(\textbf{u}) \in \{1, -1\} \\ 0 & \textrm{ otherwise } \end{array}\right.
% %255 \textrm{ for } P_{e}(\textbf{u}) \in \{1, -1\} %\left\{ \begin{array}{lr} 1 & for \ t-\Sigma_{e}(\textbf{u})\in(0;\tau) \\ 0 & for \ t-\Sigma_{e}(\textbf{u})\in(\tau;+\infty) \end{array}\right.
% \end{equation}

\subsection{Event frame}
\label{ssec:event_frame}
The second approach is a~slight modification of the previous one, as the polarity of events is taken into account, according to the work \cite{Afshar2020}: $f(\textbf{u},t) = P_{e}(\textbf{u})$.
This is one of the most popular representations due to its simplicity, but also the considerable amount of information it contains.

%PL
% Drugie podejście jest niewielką modyfikacją poprzedniego, gdyż uwzględniana jest polaryzacja zdarzeń, zgodnie z pracą \cite{Afshar2020}, co przedstawione zostało w równaniu (\ref{eq:event_frame}).
% Jest to jedna z najbardziej popularnych reprezentacji z uwagi na prostotę, ale także zawartą w sobie sporą ilość informacji.

% \begin{equation}
% \label{eq:event_frame}
% f(\textbf{u},t) = 
% %\left\{ \begin{array}{lr} 255 & \textrm{ for }  P_{e}(\textbf{u}) = 1 \\ 0 & \textrm{ for }  P_{e}(\textbf{u}) = -1 \\ 128 & \textrm{ otherwise } \end{array}\right.
% P_{e}(\textbf{u}) %\textrm{ for } P_{e}(\textbf{u}) \in \{1, -1\}
% \end{equation}

\subsection{Exponentially decaying time surface}
\label{ssec:exp_decaying_ts}
In this representation the temporal information is taken into account, precisely by specifying the time from the event occurrence to the end of the accumulation period, as in \cite{Afshar2020}.
This value is then used for determining the brightness of a~pixel in the resulting image: $f(\textbf{u},t) = P_{e}(\textbf{u}) \cdot e^{\frac{\Sigma_{e}(u)-t}{\tau}}$.
This is motivated by the observation that the weight of information contributed by an event decreases with time to 0.

%PL
% W kolejnej reprezentacji uwzględnia się już informację czasową, dokładnie poprzez określenie czasu od wystąpienia danego zdarzenia od końca okresu akumulacji, tak jak w pracy \cite{Afshar2020}.
% Wartość ta jest następnie używana podczas określania jasności piksela na obrazie wynikowym, jak w (\ref{eq:decay}).
% Jest to umotywowane spostrzeżeniem, że waga informacji wnoszonej przez dane zdarzenie maleje wraz z upływem czasu do 0.

% \begin{equation}
% \label{eq:decay}
% f(\textbf{u},t) = P_{e}(\textbf{u}) \cdot e^{\frac{\Sigma_{e}(u)-t}{\tau}} %\left\{ \begin{array}{lr} P_{e}(\textbf{u})*e^{\frac{\Sigma_{e}(\textbf{u})-t}{\tau}} & for \ \Sigma_{e}(\textbf{u}) \leq t \\ 0 & for \ \Sigma_{e}(\textbf{u}) > t \end{array}\right.
% \end{equation}

\subsection{Event frequency}
\label{ssec:event_freq}
The next approach uses information about the frequency of a~given pixel in the image, as proposed in \cite{Chen2018}: $f(x) = \frac{255}{1+e^{-x/2}}$, \added{where} the sum of polarisations for a~pixel is denoted as $x$.

%PL
% W następnym podejściu wykorzystywana jest informacja o częstotliwości występowania danego piksela na obrazie, zgodnie z wyrażeniem (\ref{eq:event_freq}), co zaproponowano w pracy \cite{Chen2018}.
% Przez $x$ oznaczona została suma polaryzacji dla danego piksela.

% \begin{equation}
% \label{eq:event_freq}
% f(x) = 255 \cdot \frac{1}{1+e^{-x/2}}
% \end{equation}

\subsection{Comparison}
\label{ssec:comp}
\begin{figure*}[!t]
    %\centering % Not needed
    \begin{subfigure}[b]{0.49\columnwidth}
        \includegraphics[width=\textwidth]{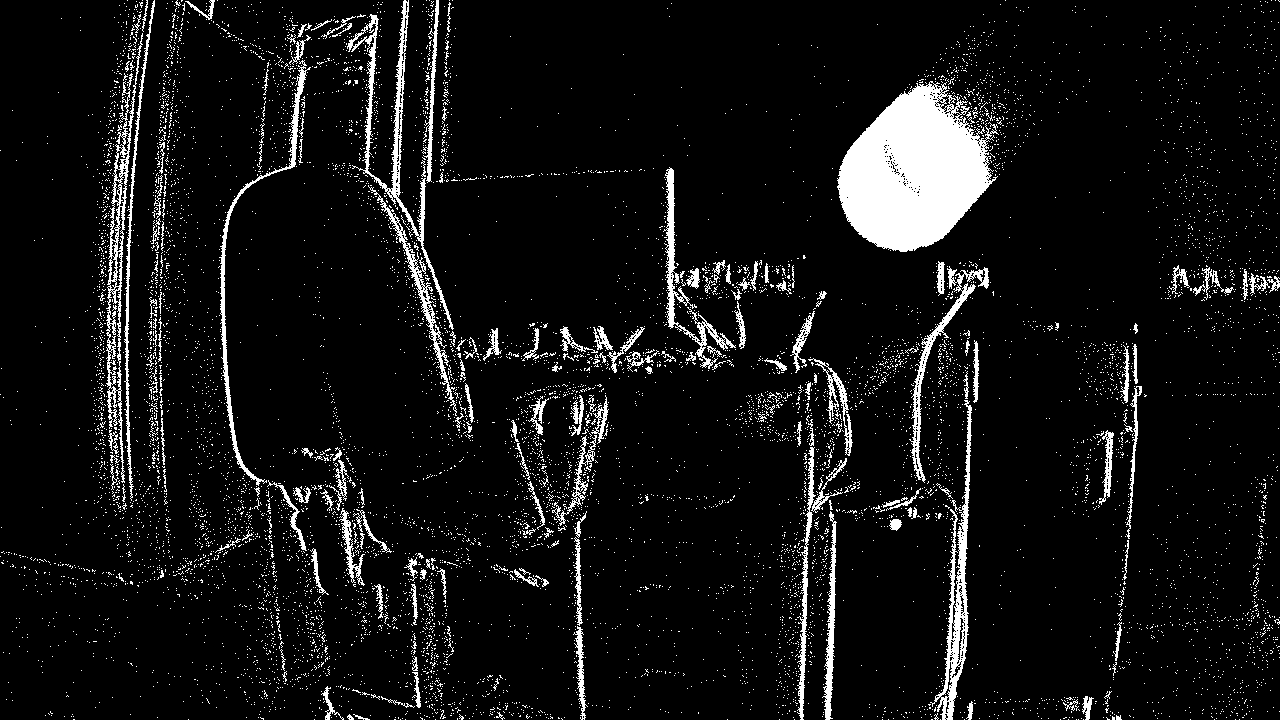}
        \caption{Binary frame}
        \label{fig:binary_frame}
    \end{subfigure}
    \hfill
    \begin{subfigure}[b]{0.49\columnwidth}
        \includegraphics[width=\textwidth]{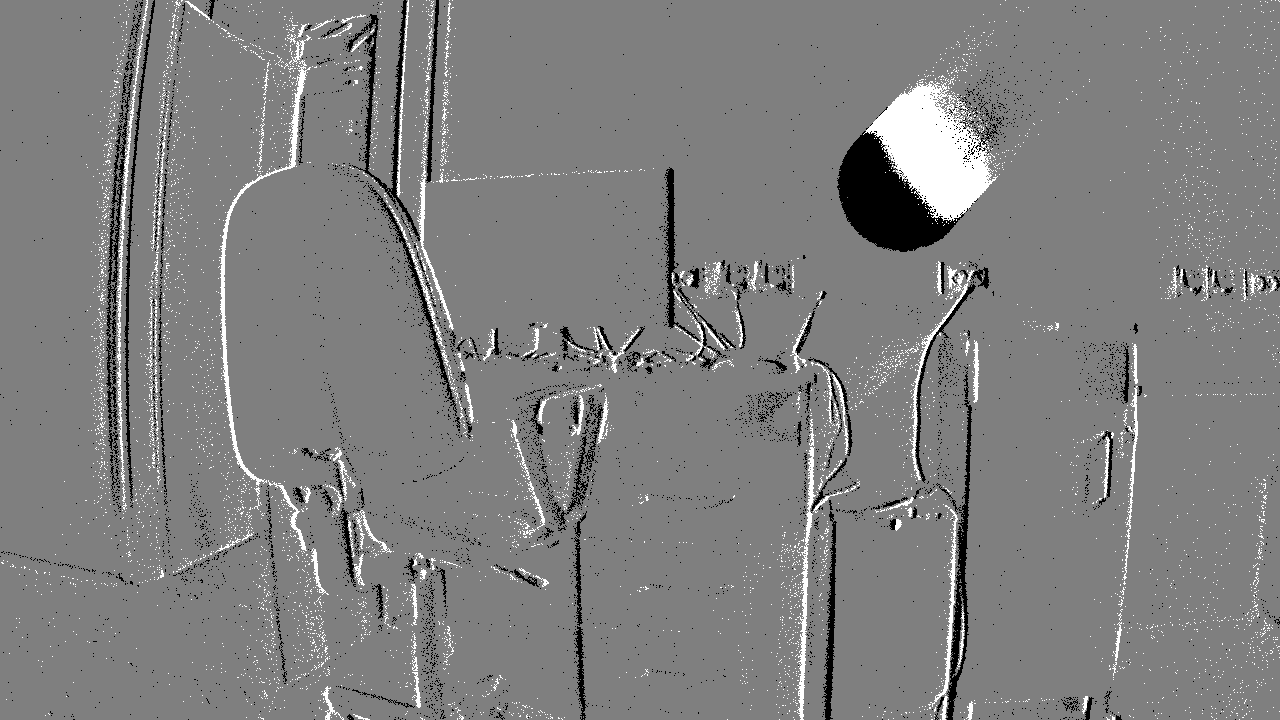}
        \caption{Event frame}
        \label{fig:event_frame}
    \end{subfigure}
    %% leave a blank line to create a line break    
    \begin{subfigure}[b]{0.49\columnwidth}
        \includegraphics[width=\textwidth]{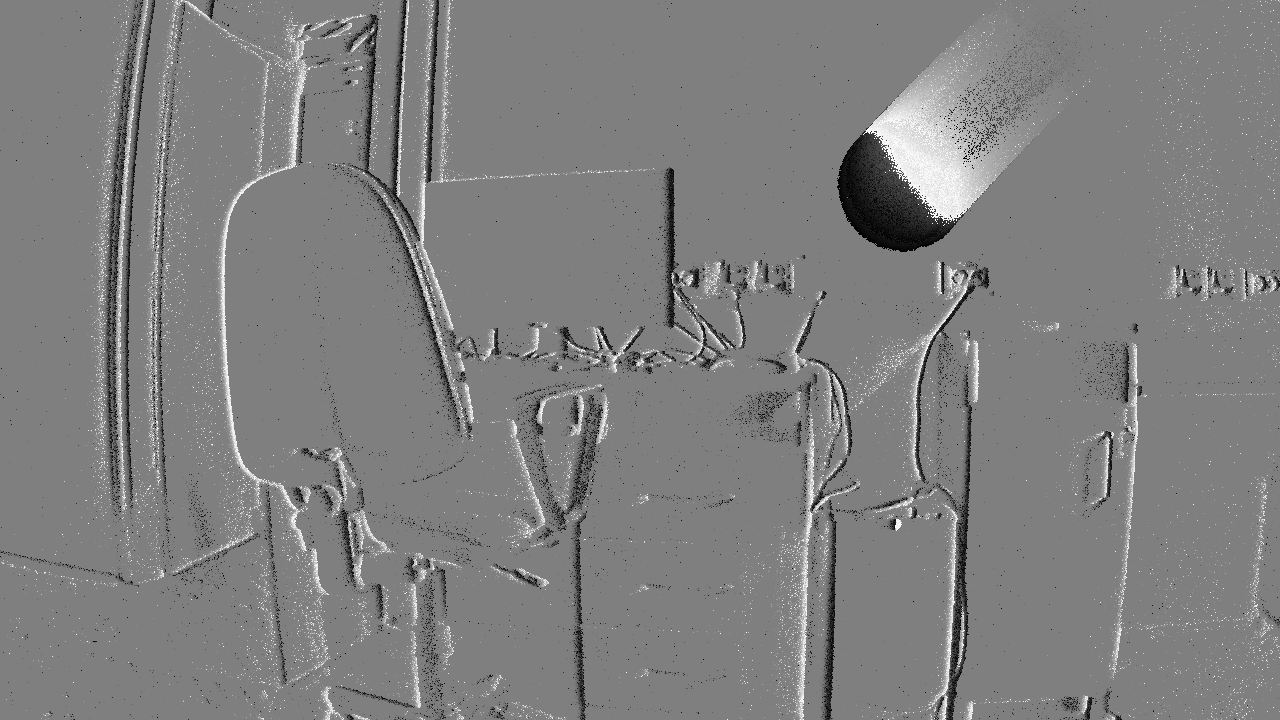}
        \caption{Exp. decaying time surface}
        \label{fig:exp_decay_ts}
    \end{subfigure}
    \hfill
    \begin{subfigure}[b]{0.49\columnwidth}
        \includegraphics[width=\textwidth]{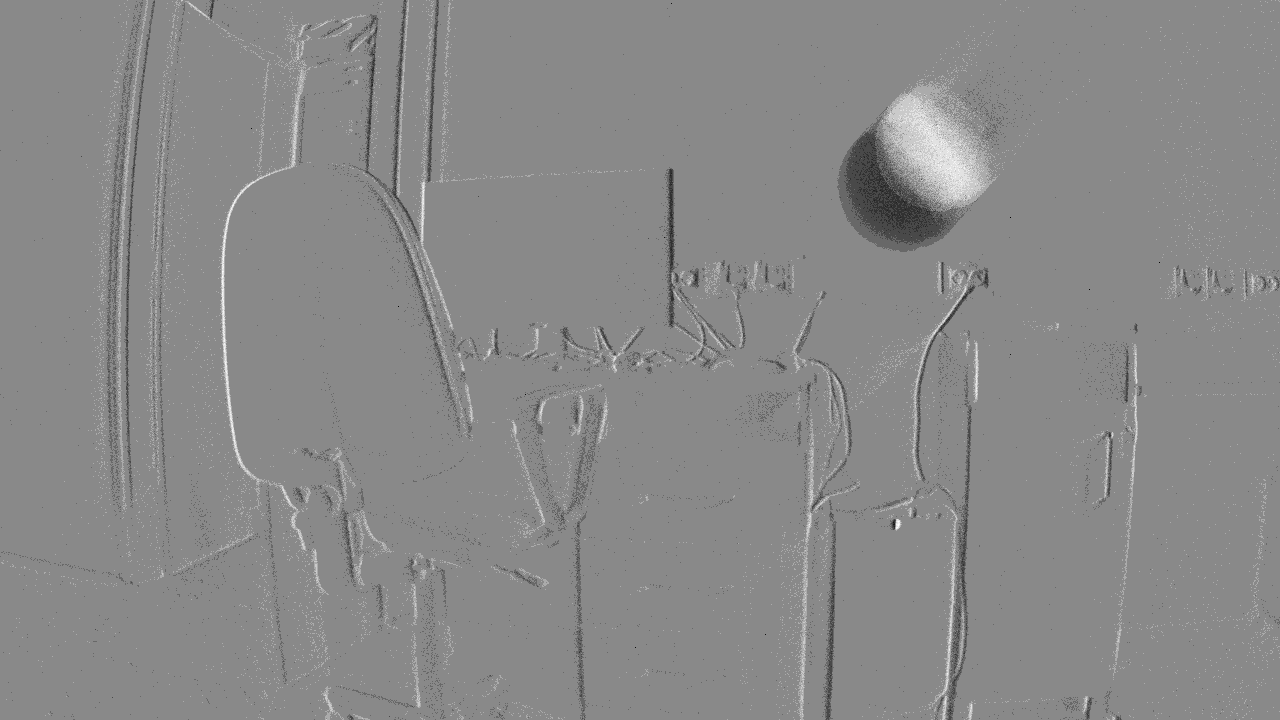}
        \caption{Event frequency}
        \label{fig:event_freq}
    \end{subfigure}
    \caption{Comparison of event frames generated using different representations over a~10 ms interval. \added{The fast-moving object generates a~big amount of events. In case of images (c) and (d), it is blurred as the influence of particular events on the resulting image differs depending on exact time or frequency.}}
    %\caption{Porównanie ramek zdarzeniowych wygenerowanych na podstawie różnych reprezentacji w przedziale 10 ms.}
    \label{fig:repr}
\end{figure*}

A~comparison of generated event frames using the representations described \deleted{for an example sequence} is shown in Figure \ref{fig:repr}.
\added{An exemplary sequence shows a~fast-moving ball flying through a~room recorded by a~moving event camera.}
Apart from aforementioned representations, more complex arithmetic operations can be proposed to determine the values of individual pixels, as \replaced{mentioned}{discussed} in Section \ref{ssec:prev_work}.
There are also solutions in which an image with multiple channels of particular representations \cite{Wzorek2022} \added{or aggregated into one-channel \cite{Innocenti2021}} is generated. %, thus exploiting the \added{additional} advantages. \deleted{ of several of them simultaneously.}

%PL
% Porównanie wygenerowanych ramek zdarzeniowych przy wykorzystaniu opisanych reprezentacji dla przykładowej sekwencji przedstawione zostało na rys. \ref{fig:repr}.
% Oprócz wymienionych reprezentacji, można zaproponować bardziej skomplikowane operacje arytmetyczne do wyznaczenia wartości poszczególnych pikseli, co omówiono w Rozdziale \ref{ssec:prev_work}.
% Istnieją także rozwiązania, w których generowany jest obraz o wielu kanałach, a poszczególne jego składowe stanowią wybrane reprezentacje \cite{Wzorek2022}, co pozwala wykorzystać zalety kilku z nich jednocześnie.

\section{The proposed frame generation module}
\label{sec:impl}
%TODO Tu coś w stylu: W ramach niniejszych prac opracowaliśmy kilka wersji modułu sprzętowego do genereacji ramek zdarzeniowych. Został on opisane w j. System Verilog i środowsku Vivado oraz przetestowany dla platformy FPGA firmy AMD Xilinx. 

In this \replaced{work}{project}, we have developed several versions of a~hardware module for event frame generation on an FPGA platform.
Their architectures were prepared in SystemVerilog language using the Vivado environment and tested for AMD Xilinx's FPGA platforms.
The input of each \added{module} was an event $e=\{t,x,y,p\}$ and the output was an 8-bit pixel brightness value.
For testing purposes, sample sequences \added{with a~thrown ball} were recorded with a~\added{moving (rotating)} Prophesee EVK1 camera in HD resolution\deleted{ ($1280 \times 720$ pixels)}.
The events were saved to a~text file, from which they were read in the Vivado simulation.

%PL
% W ramach niniejszego projektu opracowaliśmy kilka wersji sprzętowego modułu do generacji ramek zdarzeniowych na platformie FPGA.
% Architektury tych modułów opisane zostały w języku SystemVerilog i środowisku Vivado oraz przetestowane dla platform FPGA firmy AMD Xilinx.
% %w języku SystemVerilog i środowisku Vivado opracowany został moduł do generacji ramek zdarzeniowych na platformie FPGA.
% Wejście każdego z nich stanowiły zdarzenia $e={t,x,y,p}$, a wyjściem były wartości (jasności) piksela zapisane na 8 bitach.
% Na potrzeby testów nagrano przykładowe sekwencje kamerą Prophesee EVK1 w rozdzielczości HD ($1280 \times 720$ pikseli).
% Zarejestrowane zdarzenia zostały zapisane do pliku tekstowego, z którego były odczytywane w symulacji w Vivado.

\subsection{Basic version}
\label{ssec:basic}
The architecture of the basic version of the algorithm consists of three elements: a~memory for storing accumulated events, logic controlling the reading and writing of data, and a~temporary buffer for incoming data during frame reading.

The first of these elements can be realised in two ways -- using the FPGA chip's internal block memory (BRAM) or external RAM (usually dynamic).
In the first case, the memory resources are quite limited, so the appropriate representation and resolution of the image must be selected.
In the second case, the resources are much larger, but this approach increases the complexity of the architecture, introduces additional latency in data processing and is more energy consuming.
Therefore, the first solution is used in this work due to its greater simplicity and operational efficiency.
The number of elements in memory was set to 921600 (the number of pixels in an HD image)\deleted{ ($1280 \times 720$)}.
The number of bits \replaced{allocated for}{dedicated to} one element depends on the chosen representation, but in the simplest case just 1 bit can be assumed (Section \ref{ssec:binary}).
Other variants are described as algorithm extensions in Section \ref{sssec:other_repr}.

The second element is the logic that controls the communication with the two-port block memory, of which one port is used for writing data and the other for reading.
As long as events belonging to a~certain time interval $\tau$ (e.g. 10 ms) occur at the input, the algorithm operates in write mode.
Based on the coordinates of the incoming event, the address in memory is determined as: $address = address\_Y \cdot image\_width + address\_X$, while the value stored equals 1.

\begin{figure}[!t]
  \centering
  \includegraphics[width=0.5\textwidth]{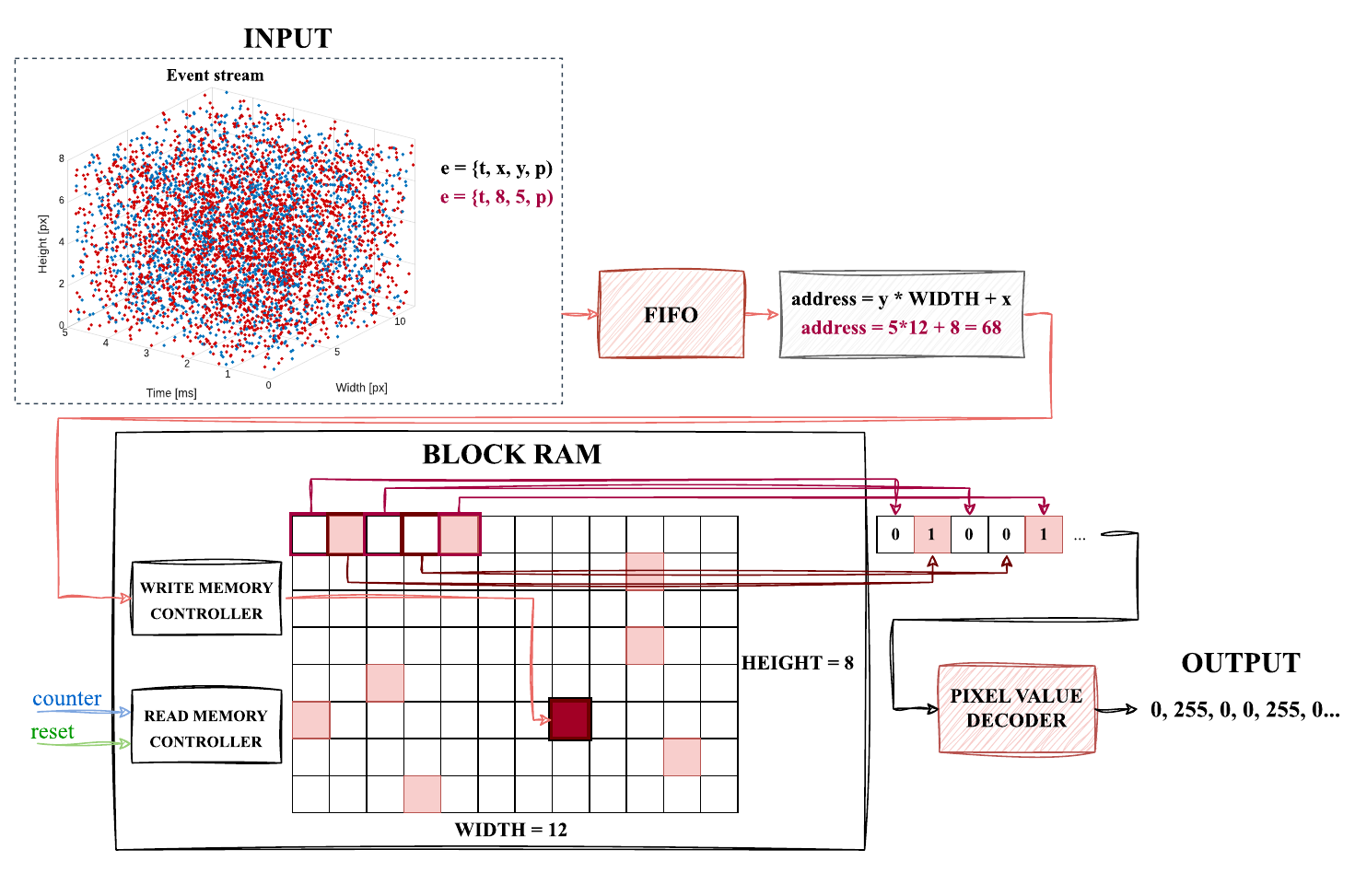}
  \caption{Operation of the frame generation algorithm in write and read mode. For simplicity, an image of $12 \times 8$ pixels was assumed. Writing was visualised using an example event $e=\{t,8,5,p\}$. Filled fields indicate earlier events. In read mode, the counter and the reset signal go through the entire image to the last pixel. Based on the representation, the values written and read from memory (and after decoding) may differ.}
  \label{fig:algo_basic_full}
\end{figure}

When the condition to stop the write mode is met (e.g. $\tau$ value exceeded), the algorithm enters read mode, resetting the pixel position counter in the image.
Its value is given as the address of the memory cell from under which the content of the accumulator is read before being incremented by 1.
The value read is further decoded to the range 0-255 to display the output event frame in greyscale.
At the same time, the address used for reading data is delayed by one clock cycle and fed to the input of the second port used for writing, together with the value 0, to reset the memory cell.
When the counter reaches the last pixel in the image, the algorithm returns to write mode.
A~schematic visualisation of the operation of the algorithm in write and read mode is shown in Figure \ref{fig:algo_basic_full}.

Due to event camera properties, just after entering the read mode, new events may appear at the input of the module, \replaced{which should}{ to} be used for the generation of the next frame.
So the last necessary element is to handle these events.
In order to separate the `older' events (in the accumulator) from the `newer' ones (at the input), a~temporary buffer must be prepared to store them until the read mode \replaced{finishes}{completed}.
It was realised as a~FIFO (First In First Out) queue using the FPGA's block memory, in which all event components were stored.

Regardless of the operation mode, each event was firstly written to the FIFO.
Once in write mode, data was read from the queue and, in the meantime, the latest events from the input were written to the FIFO.
A~difficult issue is to determine the maximum size of the queue in order to store all incoming events but with low memory usage, as it depends on the dynamics of the scene and the $\tau$ value.
In our solution, the queue size was set to 32768 elements.
In addition, a~mechanism of removing `the oldest' events from the queue to add `the newest' was implemented, if the FIFO was full in read mode.
This was motivated by the fact that `newer' events are more important, as they represent the most recent changes.

\subsection{Extensions}
\label{ssec:ext}
The architecture described in Section \ref{ssec:basic} is sufficient to correctly generate event frames in real-time on an FPGA platform.
However, a~number of enhancements and extensions can be applied to achieve better performance or results.

%PL
% Opisana w Rozdziale \ref{ssec:basic} architektura jest wystarczająca do poprawnego generowania ramek zdarzeniowych w czasie rzeczywistym na platformie FPGA.
% Można jednak zastosować szereg usprawnień i rozszerzeń, dzięki którym możliwe jest uzyskanie lepszych parametrów i wyników działania.

\subsubsection{Other representations}
\label{sssec:other_repr}
The first element is another representation of event data.
The simplest and most common modification is to add polarisation, creating an event frame.
In this case, an event with a~polarity of 1 generates an output pixel value of 255, the one with $-$1 polarity generates 0, while pixels without events receive a~value of 128.
To save hardware resources, a~2-bit value is stored in the accumulator, denoting \deleted{the number} 1 \added{(positive event)}, $-$1 \added{(negative event) or 0 (no event)}. \deleted{which is decoded to the range 0-255 after being read.}

For other representations described in Section \ref{sec:repr}, Look-Up-Table (LUT) can be used to store the approximate values of the exponential function.
In the case of the method in Section \ref{ssec:exp_decaying_ts}, knowing the value of $\tau$ (e.g. 10 ms), the difference between the current timestamp and the maximum value for the interval can be calculated.
The rounded result can be stored in an 8-bit accumulator, thus obtaining the output pixel value.

For the representation from Section \ref{ssec:event_freq}, it was necessary to determine the maximum number of events for the given coordinates, which depends on dynamics of the scene and $\tau$ value.
After tests on an example sequence, a~limitation to the interval $(-16, 15)$ turned out to be sufficient, as larger polarity sums had no apparent effect on the final pixel value.
Therefore, 5 bits were stored in the accumulator and then were decoded to an 8-bit output.
The polarity counting itself was done in a~way that with the arrival of a~new event with the given coordinates, the current value was read from the accumulator using one port of the block memory, after which, depending on the polarity, the number 1 was added or subtracted and then written back to the same address using the other port.

\subsubsection{Multiple Block RAMs}
\label{sssec:multiple_brams}

\begin{figure}[!t]
  \centering
  \includegraphics[width=0.5\textwidth]{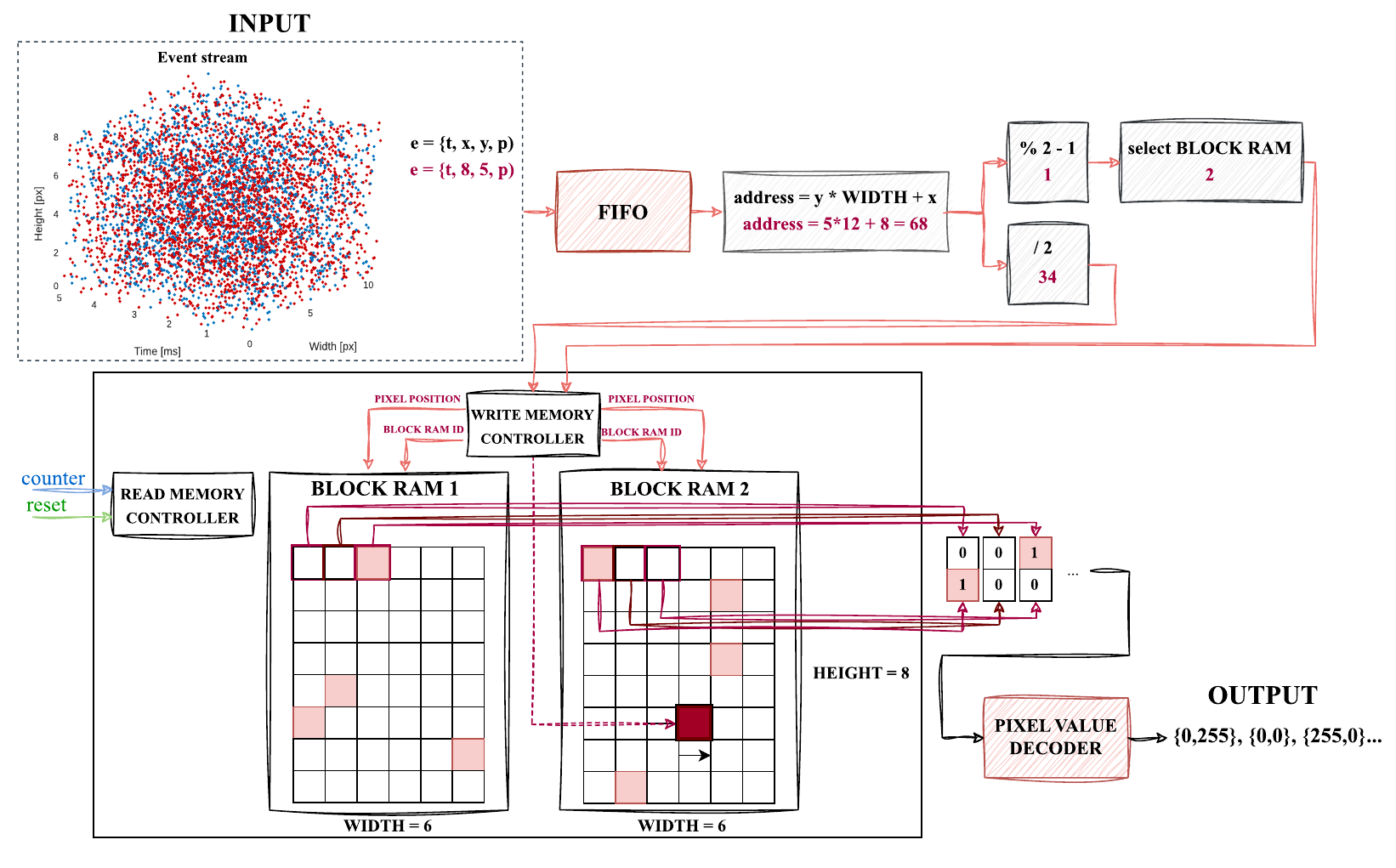}
  \caption{Use of multiple block memories for frame generation. For clarity, only 2 blocks are used in the diagram. To determine block and cell indexes, the actual number of blocks (2) is used during division. This results in a~2ppc (2 pixels per clock cycle) stream at the output of the module.}
  %\caption{Wykorzystanie wielu pamięci blokowych do generacji ramki. Dla uproszczenia liczba bloków pamięci na schemacie została ograniczona do 2. Jednak idea działania w przypadku większej liczby bloków pozostaje taka sama. Do wyznaczenia indeksu bloku pamięci oraz indeksu komórki wykorzystuje się podczas dzielenia faktyczną liczbę bloków (na zwizualizowanym przykładzie wynosi ona 2). W efekcie na wyjściu z modułu otrzymywany jest strumień pikseli w formacie 2ppc (2 pixels per clock cycle).}
  \label{fig:bram}
\end{figure}

It is possible to use a~set of smaller block memories, working in parallel, so that the reading of several pixels can be performed in the same clock cycle.
This approach is inspired by processing a~traditional video stream on an FPGA platform in a~vector format, the so-called Xppc (X pixels per clock cycle) as in \cite{Kowalczyk2018}. 
Apart from calculating the address based on the event coordinates, the horizontal index determines the memory block into which it is written, as in Figure \ref{fig:bram}.
In read mode, pixels with consecutive indices (from 0 onwards) are read from all X blocks simultaneously and fed to the module output in successive clock cycles.
With this approach, the output image is identical to that of a~single memory block, but it allows to reduce \added{X times} the latency of reading the image or \deleted{to reduce X times} the clock frequency and thus the energy consumed.
It is also possible to read multiple values from one block using memory properties in Vivado, but this solution has limited versatility in terms of possible X values and the way of resetting memory (all cells at once).

%PL
% Jednym z przetestowanych rozwiązań jest zastosowanie zestawu mniejszych pamięci blokowych, pracujących równolegle, dzięki czemu można wykonać odczyt kilku pikseli w tym samym takcie zegara.
% Podejście to jest zainspirowane przetwarzaniem tradycyjnego strumienia wideo na platformie FPGA w formacie wektorowym, tzw. Xppc (X pixels per clock cycle) jak w pracy \cite{Kowalczyk2018}. 
% Oprócz ustalenia adresu na podstawie współrzędnych zdarzenia, jego poziomy indeks określa dodatkowo blok pamięci, do którego ma zostać wpisany, tak jak w przykładzie na rys. \ref{fig:bram}.
% W trybie odczytu piksele o kolejnych indeksach (od 0) odczytywane są ze wszystkich X bloków równocześnie i podawane na wyjście modułu w kolejnych taktach zegara.
% Przy takim podejściu wyjściowy obraz jest identyczny jak przy jednym bloku pamięci, ale pozwala ono zmniejszyć latencję odczytu obrazu lub ograniczyć X-krotnie częstotliwość taktowania zegara, a tym samym zużywaną energię.

\subsubsection{Rolling window}
\label{sssec:rolling}
A~so-called rolling window, used in typical contextual operations on the image, can be proposed for event frame generation, which is partly inspired by \cite{Liu2017}.
The idea is to accumulate event data from $N$ ms, with an image generated every $K$ ms and covering the last $M$ ms, where $K \le M \le N$.
For the test, the values chosen were $N$ = 8, $M$ = 4, $K$ = 1, so a~new frame was generated every 1 ms, covering the last 4 ms, while accumulating events from the last 8 ms.
In this way, in images generated at high frequency, the oldest events are removed and the newest are added. 

\begin{figure}[!t]
  \centering
  \includegraphics[width=0.5\textwidth]{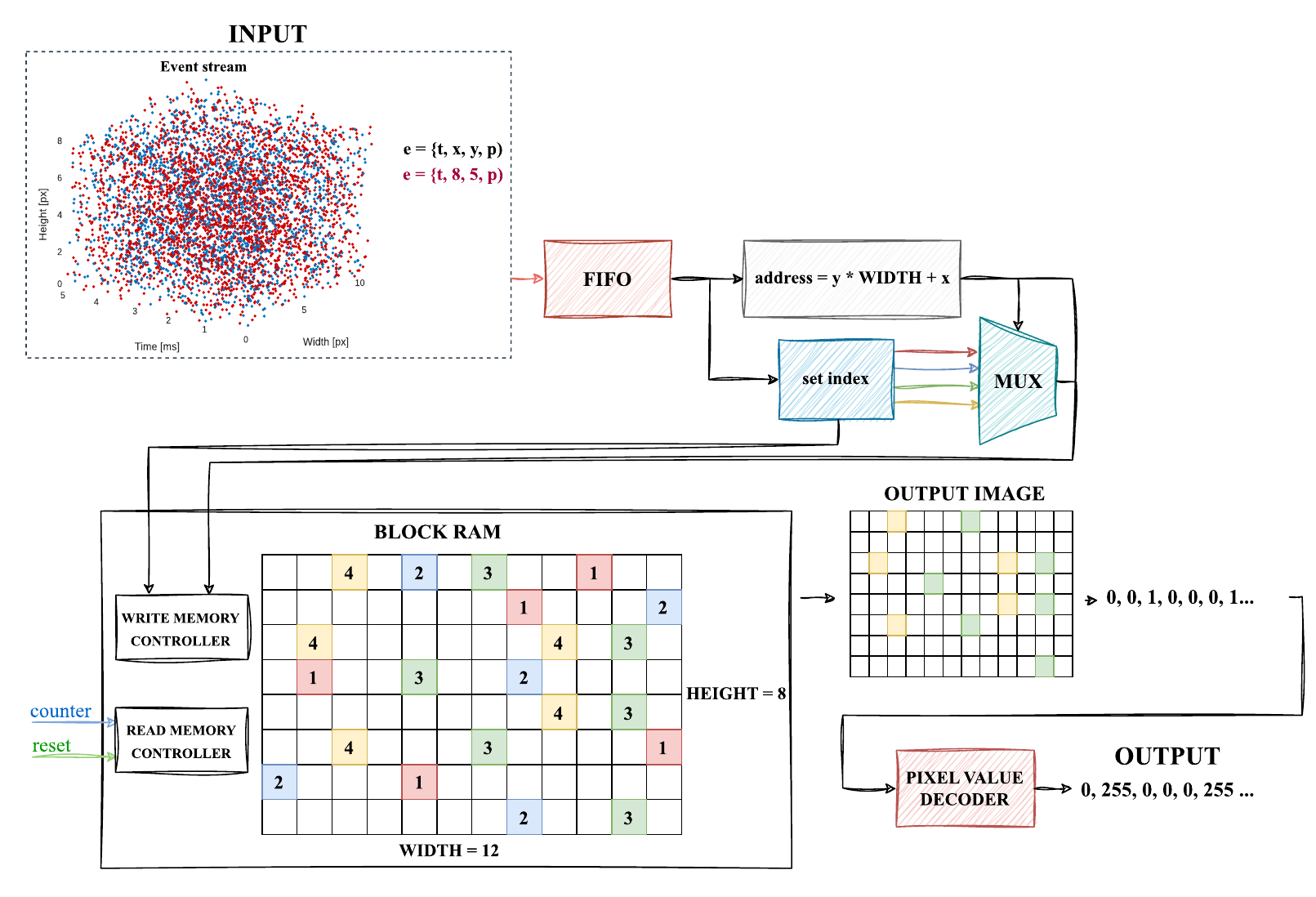}
  \caption{Operation of the rolling window method. For simplicity of the scheme, the values set are $N$ = 4, $M$ = 2, $K$ = 1. When writing to memory, apart from the address, the index of the `sub-window' is determined, in this case from 1 to 4. On reading, the indexes of the stored events are compared with the recent one and some pixels are selected and passed to the output, creating an image. In the example, only pixels from `sub-windows' 3 and 4 (green and yellow) form the output image and pixels with index 1 are removed from memory. In the next cycle, the pixels with indexes 4 and 1 will be read, and those with index 2 will be reset and so on.
  }
  %\caption{Działanie metody rolling window. Dla uproszczenia schematu przyjęto wartości $N$ = 4, $M$ = 2, $K$ = 1. Przy zapisywaniu do pamięci określany jest indeks "podokna", w tym przypadku od 1 do 4. Następnie jest wpisywany do pamięci pod odpowiedni adres wraz z wybraną reprezentacją zdarzenia. Podczas odczytu porównywane są indeksy zapisanych w akumulatorze zdarzeń wraz z indeksem aktualnym (najnowszym). Na tej podstawie wybierane są piksele, których wartości przekazywane są na wyjście, tworząc obraz, zawierający jedynie część zdarzeń przechowywanych w akumulatorze. W sytuacji przedstawionej na schemacie, tylko piksele z "podokien" 3 i 4, oznaczone dodatkowo kolorem zielonym i żółtym, są przekazywane na wyjście z modułu. Jednocześnie piksele o indeksie 1 są usuwane z pamięci, aby najnowsze, które dopiero przyjdą mogły otrzymać ten indeks i trafić do pamięci. Wtedy odczytywane będą piksele o indeksach 4 i 1, a resetowane o indeksach 2 itd.}
  \label{fig:rolling_window}
\end{figure}

On the hardware implementation side, it required additional memory due to `adding' of $\log_2 N = 3$ bits to the data representation, informing about the index of the `sub-window'.
Its resetting included only the oldest index, i.e. events before $N -$1 = 7 ms relative to the current timestamp.
A~schematic of how this method works is presented in Figure~\ref{fig:rolling_window}.

It is worth mentioning that this approach preserves more accurate information about the timestamp of events and can be applied to any data representation.
Other variants of this method may include the use of several frames, offset by $K$ ms, as components of a~multi-channel image or aggregated into a~single one as in \cite{Innocenti2021}, but also \added{the use of} dynamically changing $M$ value from which the output images are generated.

%PL
% Innym podejściem jest zaproponowanie tzw. rolling window, stosowanego w typowych operacjach kontekstowych na obrazie.
% Jego idea w przypadku generacji ramki zdarzeniowej, częściowo wzorowana na \cite{Liu2017}, polega na akumulacji danych zdarzeniowych z $N$ ms, przy czym obraz generowany jest co $K$ ms i obejmuje ostatnie $M$ ms, gdzie $K \le M \le N$.
% W ramach testów wybrano wartości $N$ = 8, $M$ = 4, $K$ = 1, zatem co 1 ms generowana była nowa ramka zdarzeniowa, obejmująca ostatnie 4 ms, natomiast w pamięci zachowywane były zdarzenia z ostatnich 8 ms.
% W ten sposób na obrazach generowanych z dużą częstotliwością usuwana jest część najstarszych zdarzeń, a w ich miejsce dodawane są najnowsze. 

% Od strony implementacji sprzętowej wymagało to dodatkowej pamięci z uwagi na "doklejanie" $\log_2 N$ = 3 dodatkowych bitów do wybranej reprezentacji danych, informujących o indeksie "podobrazu", a jej resetowanie obejmowało tylko najstarszy indeks, czyli zdarzenia sprzed $N$ - 1 = 7 ms względem aktualnego znacznika czasowego.
% Schemat działania tej metody zaprezentowany został na rys. \label{fig:rolling_window}.

% Warto dodać, że takie podejście pozwala na zachowanie dokładniejszych informacji o czasie wystąpienia zdarzeń i można zastosować je do dowolnej reprezentacji danych.
% Inne warianty tej metody mogą obejmować wykorzystanie kilku ramek, przesuniętych względem siebie o $K$ ms, jako składowych wielokanałowego obrazu wyjściowego lub zagregowanych w jeden \cite{Innocenti2021}, ale także dynamiczną zmianę wartości $M$, na podstawie której generowane są wyjściowe obrazy.

\subsubsection{Ultra RAM}
\label{sssec:ultra_ram}
Another option could be to use Ultra RAM resources available in some SoC FPGAs.
This additional internal memory is larger than BRAM and can be used to store the contents of an accumulator or a~FIFO queue. 
This can allow the use of more complex representations, but also support more dynamic scenes due to possible larger queue size.
However, only a~subset of AMD Xilinx's chips are equipped with this memory (UltraScale and UltraScale+ series), so this enhancement is dependent on the chosen hardware platform.

\subsubsection{Temporary buffer}
\label{sssec:temp_buffer}
Apart from described buffering (Section \ref{ssec:basic}), other ways of handling input events can be used.

One is to duplicate the accumulator module and use a~`ping-pong buffering' method, as in the work of \cite{Linares2021}, and swap the roles of two accumulators in a~way that one is in read mode and the other in write mode.
However, this solution is not very efficient, as the entire accumulator memory must be doubled, which may not be feasible for more complex representations due to the limited memory resources of the FPGA device.

Second idea is to use external RAM, located on the FPGA board. 
Its size is considerably larger than block memory, so a~large FIFO queue can be generated.
%An example comparison of the available memory types in \added{several popular} FPGA \added{platforms} from \added{different} AMD Xilinx \added{board families} is provided in Table \ref{table:memory}.
However, this approach also has several disadvantages, among which are: greater system complexity, higher resource utilisation, additional latency (due to reading and transferring data from memory), and the need to use the processor for communication and thus higher power consumption.
An example comparison of the available memory types in \added{several popular} FPGA \added{platforms} from \added{various} AMD Xilinx \added{board families} is provided in Table \ref{table:memory}.

\begin{table*}[t]
\caption{Comparison of available memory resources for exemplary AMD Xilinx platforms. Values in parentheses specify the number of `units' of a~memory type.}
%\caption{Porównanie dostępnych zasobów pamięciowych różnego rodzaju dla wybranych platform od AMD Xilinx. BRAM -- Block RAM, URAM -- Ultra RAM, Ext. RAM -- External RAM. Wartości w nawiasie oznaczają liczbę "sztuk" danego rodzaju pamięci.}
\centering
\label{table:memory}
\begin{tabularx}{0.575\textwidth}{lccc}%{@{} l *{5}{c} c @{}}
\toprule
SoC FPGA platform & Block RAM [Kb] & Ultra RAM [Mb] & External RAM [GB] \\ 
\midrule
ZCU 104 & 11 (312) & 27 (96) & 4.5 \\
Kria KV260 & 5 (144) & 18 (64) & 4 \\
Zybo Z7-20 & 5 (140) & - & 1 \\
\bottomrule
\end{tabularx}
\end{table*}

\subsubsection{Accumulation time}
\label{sssec:acc_time}
Event accumulation time $\tau$ is one of the most important parameters in the event frame generation algorithm.
A~smaller value provides more frames per second and \added{better} information about the dynamics of the scene, while a~larger value allows a~better understanding of the overall motion and reduces the impact of noise.
Therefore, a~potential improvement of the algorithm in this case could be an adaptive approach in which the accumulation time $\tau$ can be increased or decreased depending on the dynamics of the scene.

\subsubsection{Accumulation number}
\label{sssec:acc_number}
Another enhancement could be the generation of frames every fixed number of events $Z$, as in the work of \cite{Linares2021}.
Event data is recorded asynchronously for each pixel and the dynamics of the scene determines the exact number of events per time interval.
In this approach, it may possible (depending on the representation) to omit the timestamps and reduce the amount of data processed.
However, the generated frames will appear at the output in different time intervals, which may raise problems especially in cases with high dynamics \added{(and thus high fps value)}.
A~potential solution to this problem could be an adaptive version, in which the length of the time interval is additionally analysed and the number of $Z$ events is reduced or increased based on it.

%PL
% Inne usprawnienie algorytmu stanowić może generacja ramek co ustaloną liczbę zdarzeń $Z$, jak w pracy \cite{Linares2021}.
% Dane te rejestrowane są asynchronicznie dla każdego piksela i to od dynamiki sceny zależy ich dokładna liczba w przedziale czasu.
% W takim podejściu możliwe jest (w zależności od przyjętej reprezentacji) pominięcie znaczników czasowych i zmniejszenie ilości przetwarzanych danych.
% Z drugiej strony, generowane ramki będą się pojawiać na wyjściu w różnych odstępach czasu, co może rodzić problemy zwłaszcza w przypadkach o dużej dynamice.
% Potencjalnym rozwiązaniem tego problemu może być wersja adaptacyjna, w której dodatkowo analizowana jest długość przedziału czasowego i na tej podstawie liczba zdarzeń $Z$ jest odpowiednio zmniejszana lub zwiększana.

\subsection{Performance}
\label{ssec:perf}

\begin{table*}[t]
\caption{Comparison of hardware resource utilisation on an FPGA platform for different data representations and variants (for a~2-bit event frame only) for a~10 ms accumulation interval. The size of the FIFO queue was set to the minimum (512 elements) not to disturb the comparisons. Power estimation \added{for the FPGA chip only} was performed by the Vivado software. Due to the nature of BRAM memory, the smallest indivisible element has 0.5 size of a~single block (meant as a~`piece').}
%\caption{Porównanie wykorzystania zasobów sprzętowych na platformie FPGA dla różnych reprezentacji dla czasu akumulacji zdarzeń równego 10 ms. Rozmiar kolejki FIFO został ustawiony na minimalny (512 elementów), aby nie zaburzać porównania samych reprezentacji. Estymacja mocy jest wykonywana przez program Vivado. Z uwagi na specyfikę pamięci BRAM, najmniejszy niepodzielny element ma rozmiar równy 0.5 pojedynczego bloku ("sztuki").}
\centering
\label{tab:repr}
\begin{tabularx}{0.73\textwidth}{lccccc}%{@{} l *{5}{c} c @{}}
\toprule
Representation/Algorithm variant & No. of bits & Block RAM & LUT & Flip-Flop & Power est. [W] \\ 
\midrule
Binary frame (Sec. \ref{ssec:basic}) & 1 & 29.5 & 191 & 142 & 5.9 \\
Event frame (Sec. \ref{sssec:other_repr}) & 2 & 57.5 & 240 & 157 & 9.0 \\
Exp. decaying time surface (Sec. \ref{sssec:other_repr}) & 8 & 226 & 1289 & 168 & 29.1 \\
Event frequency (Sec. \ref{sssec:other_repr}) & 5 & 142 & 363 & 158 & 27.3 \\
\midrule
Basic (Sec. \ref{ssec:basic}) & 2 & 57.5 & 240 & 157 & 9.0 \\
Multiple BRAMs (Sec. \ref{sssec:multiple_brams}) & 2 & 61 & 204 & 258 & 12.0 \\
Rolling window (Sec. \ref{sssec:rolling}) & 5 & 142 & 8369 & 150 & 18.2 \\
\bottomrule
\end{tabularx}
\end{table*}

A~comparison of the most important parameters of several data representations and variants for event frame generation algorithm is provided in Table \ref{tab:repr}.
The use of memory resources increases significantly with the number of bits per pixel due to larger accumulator size.
Due to the amount of available memory resources, some options described may not be feasible to implement on smaller FPGA chips -- therefore simpler representations, additional approximations or lower data resolution may be needed.
The use of the remaining resources is very low, as the logic itself controlling the writing and reading of data from the accumulator is relatively simple.
%A comparison of the hardware resource consumption and the most important parameters of several variants of the developed algorithm is presented in Table \ref{tab:variants}.

The described variants concern only one most popular data representation (event frame).
In case of the rolling window method, it is necessary to allocate more memory to store the `sub-window' indices (3 extra bits -- 8 indices).
The use of multiple memory blocks generates a~slight increase in resource consumption, because with suboptimal parameters, a~part of each allocated block remains unused.
However, this solution allows faster reading of pixels, vector data processing (Section \ref{sssec:multiple_brams}) and reduced clock frequency and power consumption.
Therefore, using one block configured in Vivado with a~specified set of parameters and a~global reset can also be considered, if smaller versatility of the module is not a~problem.

\section{Conclusion}
\label{sec:concl}
In this paper we have proposed and compared various ways of event frame generation in SoC FPGA \added{devices for} \deleted{.
In particular, we focused on} an HD \deleted{($1280 \times 720$ pixels)} event stream.
\deleted{The resulting event frames can be used in typical vision algorithms with both classical methods and deep neural networks.}
\deleted{On the other hand, the use of SoC FPGAs should yield real-time and energy-efficient event data processing systems.}
For a~pipelined hardware implementation on an FPGA platform, this operation brings a~number of challenges and trade-offs depending on the available resources, mainly memory: the choice of resolution of the generated frames, the representation used, the realisation of a~temporary buffer and additional hardware enhancements.

\added{The use of SoC FPGA chips from AMD Xilinx can yield real-time and energy-efficient data processing, but} the resources available on a~chosen platform condition the details of the implementation.
In case of Zybo Z7-20, only the simplest representations (binary and event frame) can be realised due to the small size of the available memory.
For Kria KV260, any of the presented variants can be used, but Ultra RAM or external memory must be used, while for ZCU 104, all of them can be implemented using only block memory.

As part of future work, several applications of the proposed event frame generation module are planned, \added{as the frames themselves can be used in typical vision-based systems with both classical methods and deep neural networks.}
Running on an exemplary hardware platform together with the loading of data from an SD card \replaced{(or later receiving}{and then transmitted} directly from the camera), it can be possible to use this module in a~larger vision system, e.g. for the detection of fast-moving objects.
Ultimately, the developed system is planned to be used on an unmanned autonomous drone to enable it, among other things, to fly through a~dynamic environment with obstacles.  

\end{document}